\documentclass[letterpaper, 10 pt, conference]{ieeeconf}  % Comment this line out if you need a4paper

\IEEEoverridecommandlockouts                              % This command is only needed if 
\usepackage{amsmath} % assumes amsmath package installed
\usepackage{amssymb}  % assumes amsmath package installed
\usepackage{color}
\usepackage{svg}
\usepackage{tikz}
\usepackage{tikz-3dplot}
\usepackage{subfig}
\usepackage{siunitx}
\usepackage{url}
\usepackage{balance}

\definecolor{elia}{rgb}{0.5,0.5,0.8}

\newcommand{\rom}[1]{\uppercase\expandafter{\romannumeral #1\relax}}
\definecolor{alexey}{rgb}{0.8,0.0,0.8}

\definecolor{rene}{rgb}{0.0, 0.0, 1,0}
\let\OLDthebibliography\thebibliography
\renewcommand\thebibliography[1]{
  \OLDthebibliography{#1}
  \setlength{\parskip}{2pt}
  \setlength{\itemsep}{0pt plus 0.3ex}
}
\newcommand{\videolink}{{\footnotesize{\url{https://youtu.be/UuQvijZcUSc}}}}

\def \waypoint {gate}
\def \waypoints {gates}

\newcommand{\darkgrayed}[1]{\textcolor{darkgray}{#1}}
\makeatletter
\newcommand*\titleheader[1]{\gdef\@titleheader{#1}}
\AtBeginDocument{%
  \let\st@red@title\@title
  \def\@title{%
    \vskip-3em
    \bgroup\normalfont\large\centering\@titleheader\par\egroup
    \vskip1.5em\st@red@title}
}
\makeatother

\titleheader{\darkgrayed{This paper has been accepted for publication at the IEEE International Conference on Robotics and Automation (ICRA), Montreal, 2019.
\copyright IEEE}}

 \title{\LARGE \bf Beauty and the Beast:\\Optimal Methods Meet Learning for Drone Racing}

\author{Elia Kaufmann$^{1}$, Mathias Gehrig$^{1}$, Philipp Foehn$^{1}$,
        \\ Ren\'{e} Ranftl$^{2}$, Alexey Dosovitskiy$^{2}$, Vladlen Koltun$^{2}$, Davide Scaramuzza$^{1}$
\thanks{This work was supported by the Intel Network on Intelligent Systems, the National Centre of Competence in Research Robotics (NCCR) through the Swiss National Science Foundation and the SNSF-ERC Starting Grant.}
\thanks{$^{1}$ Robotics and Perception Group, Dep. of Informatics, University of Zurich, Dep. of Neuroinformatics, University of Zurich and ETH Zurich. }%
\thanks{$^{2}$ Intelligent Systems Lab, Intel}%
}

\begin{document}

\maketitle
\thispagestyle{empty}
\pagestyle{empty}

\begin{abstract}
Autonomous micro aerial vehicles still struggle with fast and agile maneuvers, dynamic environments, imperfect sensing, and state estimation drift. Autonomous drone racing brings these challenges to the fore. Human pilots can fly a previously unseen track after a handful of practice runs. In contrast, state-of-the-art autonomous navigation algorithms require either a precise metric map of the environment or a large amount of training data collected in the track of interest. To bridge this gap, we propose an approach that can fly a new track in a previously unseen environment without a precise map or expensive data collection. Our approach represents the global track layout with coarse gate locations, which can be easily estimated from a single demonstration flight. At test time, a convolutional network predicts the poses of the closest gates along with their uncertainty. These predictions are incorporated by an extended Kalman filter to maintain optimal maximum-a-posteriori estimates of gate locations. This allows the framework to cope with misleading high-variance estimates that could stem from poor observability or lack of visible gates. Given the estimated gate poses, we use model predictive control to quickly and accurately navigate through the track. We conduct extensive experiments in the physical world, demonstrating agile and robust flight through complex and diverse previously-unseen race tracks. The presented approach was used to win the IROS 2018 Autonomous Drone Race Competition, outracing the second-placing team by a factor of two.
\end{abstract}
\section*{Supplementary material}
\noindent Video: \videolink

\section{Introduction}
\begin{figure}
    \centering
    \includegraphics[width=\linewidth,trim={50 0 50 0},clip]{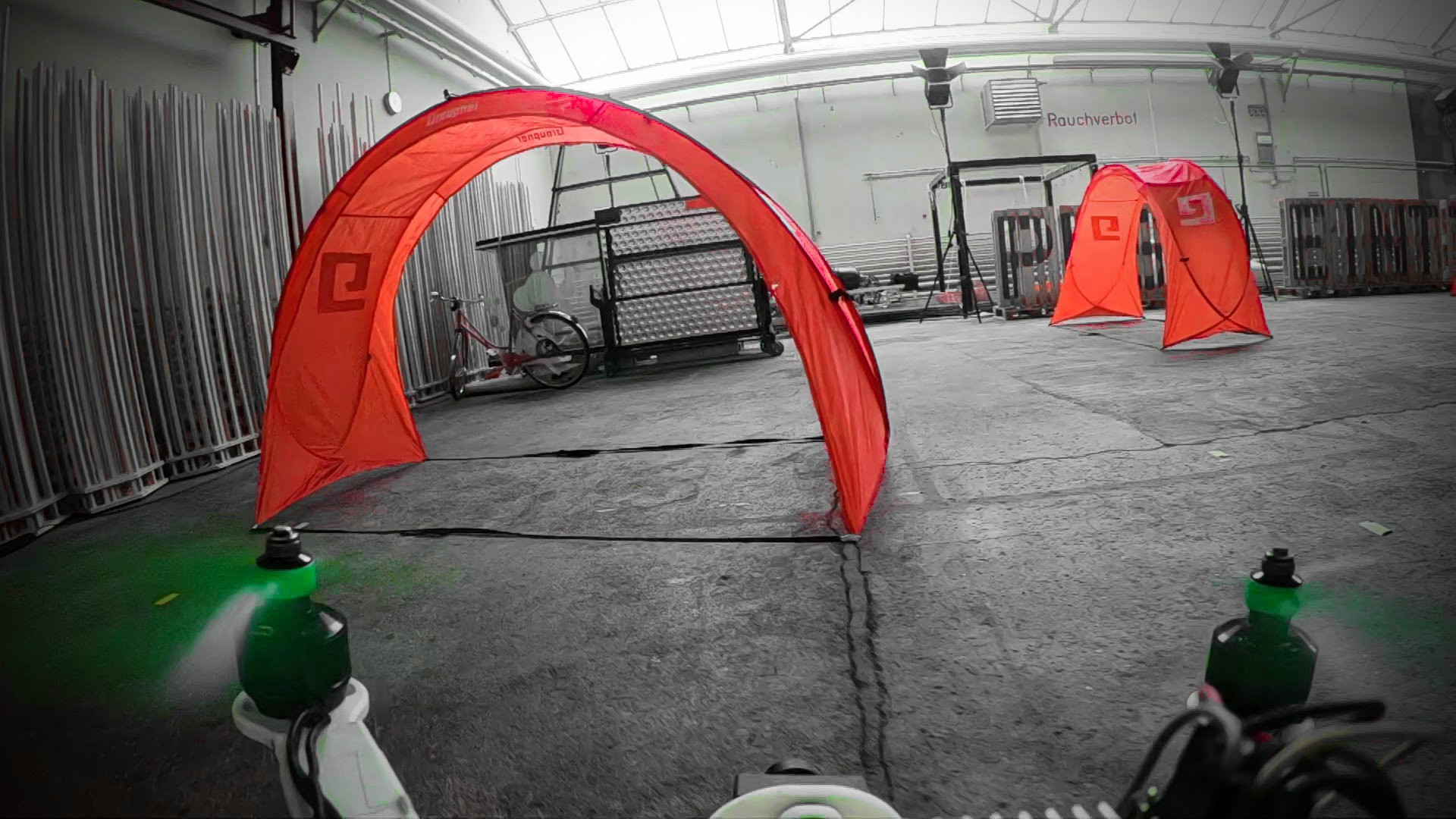}
    \caption{A quadrotor flies through an indoor track. Our approach uses optimal filtering to incorporate estimates from a deep perception system. It can race a new track after a single demonstration.}
    \label{fig:my_label}
\end{figure}

First-person view (FPV) drone racing is a fast-growing sport, in which human pilots race micro aerial vehicles (MAVs) through tracks via remote control. Drone racing provides a natural proving ground for vision-based autonomous drone navigation. This has motivated competitions such as the annual IROS Autonomous Drone Race~\cite{Moon16iros} and the recently announced AlphaPilot Innovation Challenge, an autonomous drone racing competition with more than 2 million US dollars in cash prizes.

To successfully navigate a race track, a drone has to continually sense and interpret its environment. It has to be robust to cluttered and possibly dynamic track layouts. It needs precise planning and control to support the aggressive maneuvers required to traverse a track at high speed. Drone racing thus crystallizes some of the central outstanding issues in robotics. Algorithms developed for drone racing can benefit robotics in general and can contribute to areas such as autonomous transportation, delivery, and disaster relief.

Traditional localization-based approaches for drone navigation require precomputing a precise 3D map of the environment against which the MAV is localized.
Thus, while previous works demonstrated impressive results in controlled settings~\cite{morrell2018}, these methods are difficult to deploy in new environments where a precise map is not available. Additionally, they fail in the presence of dynamic objects such as moving gates, have inconsistent computational overhead, and are prone to failure under appearance changes such as varying lighting.

Recent work has shown that deep networks can provide drones with robust perception capabilities and facilitate safe navigation even in dynamic environments~\cite{Kaufmann18arxiv,Jung18ral}. However, current deep learning approaches to autonomous drone racing require a large amount of training data collected in the same track. This stands in contrast to human pilots, who can quickly adapt to new tracks by leveraging skills acquired in the past.

In this paper, we develop a deep-learning-aided approach to autonomous drone racing capable of fast adaptation to new tracks, without the need for building precise maps or collecting large amounts of data from the track.
We represent a track by coarse locations of a set of \waypoints, which can be easily acquired in a single demonstration flight through the track.
These recorded {\waypoints} represent the rough global layout of the track. At test time, the local track configuration is estimated by a convolutional network that predicts the location of the closest {\waypoint} together with its uncertainty, given the currently observed image. 
The network predictions and uncertainties are continuously incorporated using an extended Kalman filter (EKF) to derive optimal maximum-a-posteriori estimates of {\waypoint} locations.
This allows the framework to cope with misleading high-variance estimates that could stem from bad observability or complete absence of visible {\waypoints}.
Given these estimated {\waypoint} locations, we use model predictive control to quickly and accurately navigate through them.

We evaluate the proposed method in simulation and on a real quadrotor flying fully autonomously. Our algorithm runs onboard on a computationally constrained platform. We show that the presented approach can race a new track after only a single demonstration, without any additional training or adaptation. Integration of the estimated {\waypoint} positions is crucial to the success of the method: a purely image-based reactive approach only shows non-trivial performance in the simplest tracks. We further demonstrate that the proposed method is robust to dynamic changes in the track layout induced by moving gates.

The presented approach was used to win the IROS Autonomous Drone Race Competition, held in October 2018. An MAV controlled by the presented approach placed first in the competition, traversing the eight gates of the race track in 31.8 seconds. In comparison, the second-place entry completed the track in 61 seconds, and the third in 90.1 seconds.

\section{Related Work}

Traditional approaches to autonomous MAV navigation build on visual inertial odometry (VIO) \cite{Forster17troSVO, Blosch15iros, leutenegger2015keyframe, usenko2016direct} or simultaneous localization and mapping (SLAM) \cite{MurArtal17tro, schneider2018maplab}, which are used to provide a pose estimate of the drone relative to an internal metric map  \cite{Loianno17ral,fehr2018visual}.
While these methods can be used to perform visual teach and repeat \cite{fehr2018visual}, they are not concerned with trajectory generation \cite{Mellinger11icra, Mueller13icra}.
%For tasks that are time-sensitive, the quality of trajectories is crucial \cite{Mellinger11icra, Mueller13icra}.
Furthermore, teach and repeat assumes a static world and accurate pose estimation: assumptions that are commonly violated in the real world.
%However, such assumptions are violated in the real world and might lead to deteriorated performance.

The advent of deep learning has inspired alternative solutions to autonomous navigation that aim to overcome these limitations. These approaches typically predict actions directly from images.
Output representations range from predicting discrete navigation commands (classification in action space)~\cite{kim2015deep,Giusti16ral,loquercio2018dronet} to direct regression of control signals~\cite{mueller2017teaching}.
A different line of work combines network predictions with model predictive control by regressing the cost function from a single image~\cite{drews2017aggressive}.

In the context of drone racing, Kaufmann et al.~\cite{Kaufmann18arxiv} proposed an intermediate representation in the form of a goal direction and desired speed. 
The learned policy imitates an optimal trajectory~\cite{Mellinger11icra} through the track.
An advantage of this approach is that it can navigate even when no gate is in view, by exploiting track-specific context and background information.
A downside, however, is the need for a large amount of labeled data collected directly in the track of interest in order to learn this contextual information.
As a result, the approach is difficult to deploy in new environments.

Jung et al.~\cite{Jung18ral} consider the problem of autonomous drone navigation in a previously unseen track. They use line-of-sight guidance combined with a deep-learning-based gate detector. As a consequence, the next gate to be traversed has to be in view at all times. Additionally, gates cannot be approached from an acute angle since the algorithm does not account for gate rotation. The method is thus applicable only to relatively simple environments, where the next gate is always visible.

Our approach addresses the limitations of both works~\cite{Kaufmann18arxiv,Jung18ral}. It operates reliably even when no gate is in sight, while eliminating the need to retrain the perception system for every new track.
This enables rapid deployment in complex novel tracks.

\section{Methodology}
We address the problem of robust autonomous flight through a predefined, ordered set of possibly spatially perturbed \waypoints.
Our approach comprises three subsystems: perception, mapping, and combined planning and control.
The perception system takes as input a single image from a forward-facing camera and estimates both the relative pose of the next {\waypoint} and a corresponding uncertainty measure.
The mapping system receives the output of the perception system together with the current state estimate of the quadrotor and produces filtered estimates of {\waypoint} poses. 
The {\waypoint} poses are used by the planning system to maintain a set of waypoints through the track.
These waypoints are followed by a control pipeline that generates feasible receding-horizon trajectories and tracks them.

\begin{figure}[!h]
\centering
\tdplotsetmaincoords{60}{35}
\begin{tikzpicture}[tdplot_main_coords, scale=2]

\def\framearrow{0.4}
\coordinate (e1) at (\framearrow,0,0);
\coordinate (e2) at (0,\framearrow,0);
\coordinate (e3) at (0,0,\framearrow);

\def\gatewidthhalf{0.5}
\def\gateheighthalf{0.5}

\coordinate (odometry_frame_origin) at (0, 0, -0.8);
\coordinate (body_frame_origin) at (0.7,0.4,0.4);
\coordinate (reference_frame_origin) at (0.8,0.5,0.3);
\coordinate (gate_frame_origin) at (3,1.0,0.2);

\coordinate (next_gate_corner_1) at ($ (gate_frame_origin) + (0, \gatewidthhalf, \gateheighthalf) $);
\coordinate (next_gate_corner_2) at ($ (gate_frame_origin) + (0, -\gatewidthhalf, \gateheighthalf) $);
\coordinate (next_gate_corner_3) at ($ (gate_frame_origin) + (0, -\gatewidthhalf, -\gateheighthalf) $);
\coordinate (next_gate_corner_4) at ($ (gate_frame_origin) + (0, \gatewidthhalf, -\gateheighthalf) $);

% odometry frame
\draw[thick,->,color=red,text=black] (odometry_frame_origin) -- ($ (odometry_frame_origin) + (e1)$) node[right] {$x$};		
\draw[thick,->,color=green,text=black] (odometry_frame_origin) -- ($ (odometry_frame_origin) + (e2)$) node[right] {$y$};		
\draw[thick,->,color=blue,text=black] (odometry_frame_origin) -- ($ (odometry_frame_origin) + (e3)$) node[above, left] {$z$};	
\node[draw=none] at ($(odometry_frame_origin) + (0.0, 0, -0.2)$)  {$O$};

% body frame
\draw[thick,->,color=red,text=black] (body_frame_origin) -- ($(body_frame_origin) + (e1)$);
\draw[thick,->,color=green,text=black] (body_frame_origin) -- ($(body_frame_origin) + (e2)$);
\draw[thick,->,color=blue,text=black] (body_frame_origin) -- ($(body_frame_origin) + (e3)$);
\node[draw=none] at ($(body_frame_origin) + (0.0, 0, -0.2)$)  {$B$};

% gate frame
\draw[thick,->,color=red,text=black] (gate_frame_origin) -- ($(gate_frame_origin) + (e1)$);	
\draw[thick,->,color=green,text=black] (gate_frame_origin) -- ($(gate_frame_origin) + (e2)$);
\draw[thick,->,color=blue,text=black] (gate_frame_origin) -- ($(gate_frame_origin) + (e3)$);
\node[draw=none] at ($(gate_frame_origin) + (0.0, 0, -0.2)$)  {$\text{G}_l$};

% Transformations
\draw[thin,->,color=black,text=black] (odometry_frame_origin) -- (body_frame_origin) node[below=15, left=15] {$(\mathbf{t}_{OB},\mathbf{R}_{OB})$};
\draw[thin,->,color=black,text=black] (body_frame_origin) -- (gate_frame_origin) node[above=35, left=10] {($\mathbf{t}_{BG_l},\mathbf{R}_{BG_l}),~\boldsymbol{\Sigma}_{BG_l}$};
\draw[thin,->,color=black,text=black] (odometry_frame_origin) -- (gate_frame_origin) node[below=25, left=30] {$(\mathbf{t}_{OG_l},\mathbf{R}_{OG_l}),~\boldsymbol{\Sigma}_{OG_l}$};

% next gate
\draw[thick,-,color=black,text=black]
    (next_gate_corner_1)--(next_gate_corner_2)--(next_gate_corner_3)--(next_gate_corner_4)--cycle;

\end{tikzpicture}
\caption{Relation of odometry $O$, body $B$, and {\waypoint} frame $G_l$.}
\label{fig:frames}
\vspace{-3mm}
\end{figure}
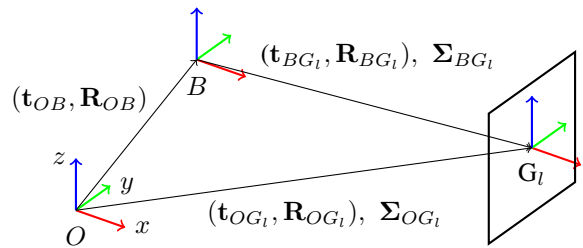
\begin{figure*}[!t]
\centering
\includegraphics[width=\linewidth] {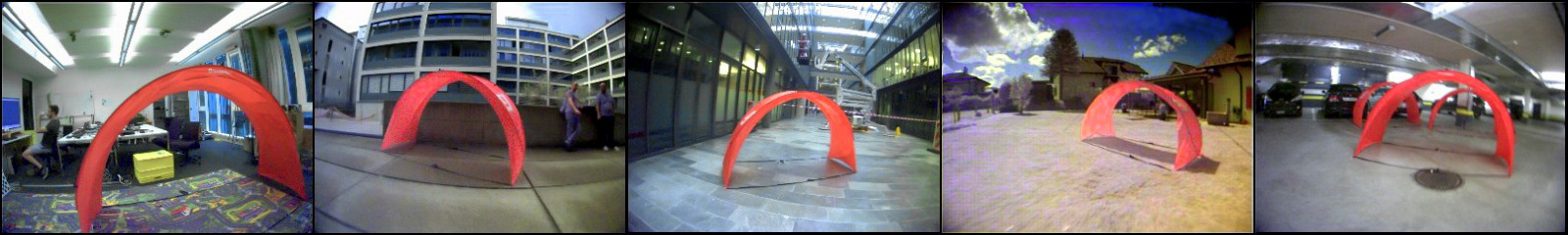}
\caption{We collected training data for the perception system in 5 different environments. From left to right: flying room, outdoor urban environment, atrium, outdoor countryside, garage.}
\label{fig:training_data}
\vspace{-3mm}
\end{figure*}

\subsection{Notation and Frame Convention}
We denote all scalars by lowercase letters $x$, vectors by lowercase bold letters $\mathbf{x}$, and matrices by bold uppercase letters $\mathbf{X}$. Estimated values are written as $\hat{x}$, measured values as $\tilde{x}$.

The relevant coordinate frames are the odometry frame $O$, the body frame $B$, and the {\waypoint} frames $G_l$, where ${l \in \{1,\ldots, N_l\}}$ and $N_l$ is the number of {\waypoints}. A schematic overview of the relation between coordinate frames is shown in Figure \ref{fig:frames}. The odometry frame $O$ is the global VIO reference frame.
The relation between the body frame $B$ and the odometry frame $O$ is given by the rotation  $\mathbf{R}_{OB}$ and translation $\mathbf{t}_{OB}$.
This transform is acquired through a visual inertial pose estimator.
The prediction $(\tilde{\mathbf{t}}_{BG_l}, \tilde{\mathbf{R}}_{BG_l})$ is provided together with a corresponding uncorrelated covariance in polar coordinates $\tilde{\boldsymbol{\Sigma}}_{BG_l,pol}=\text{diag}(\tilde{\boldsymbol{\sigma}}_{BG_l,pol}^2)$ of the {\waypoint}'s pose in the body frame.
In parallel, we maintain an estimate of each {\waypoint} pose $(\hat{\mathbf{t}}_{OG_l}, \hat{\mathbf{R}}_{OG_l})$ along with its covariance $\hat{\boldsymbol{\Sigma}}_{OG_l} = \text{cov}\left(\hat{\mathbf{t}}_{OG_l}, \hat{\mathbf{R}}_{OG_l}\right)$ in the odometry frame. This has the advantage that {\waypoint} poses can be updated independently of each other.

\subsection{Perception System}
\subsubsection{Architecture}
The deep network takes as input a ${320\times240}$ RGB image and regresses both the mean ${\tilde{\mathbf{z}}_{BG_l,pol}=[\tilde{r}, \tilde{\theta}, \tilde{\psi}, \tilde{\phi}]^\top\in \mathbb{R}^4}$ and the variance ${\tilde{\boldsymbol{\sigma}}_{BG_l,pol}^2\in \mathbb{R}^4}$ of a multivariate normal distribution that describes the current estimate of the next {\waypoint}'s pose.
Our choice of output distribution is motivated by the fact that we use an EKF to estimate the joint probability distribution of a {\waypoint}'s pose, which is known to be optimal for identical and independently distributed white noise with known covariance.
The mean represents the prediction of the relative position and orientation of the {\waypoint} with respect to the quadrotor in spherical coordinates.
We found this to be advantageous compared to a Cartesian representation since it decouples distance estimation from the position of the {\waypoints} in image coordinates.
We use a single angle $\tilde{\phi}$ to describe the relative horizontal orientation of the {\waypoint}, since the gravity direction is known from the IMU.
Furthermore, we assume that {\waypoints} are always upright and can be traversed horizontally along the normal direction.
Specifically, $\tilde{\phi}$ is measured between the quadrotor's current heading and the {\waypoint}'s heading.

\begin{figure}[htbp]
  \centering
  \input{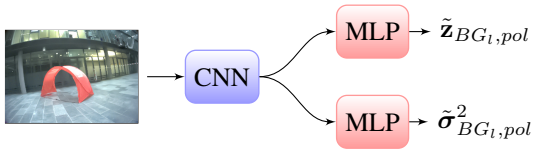}
  \caption{Schematic illustration of the network architecture. Image features are extracted by a CNN \cite{loquercio2018dronet} and passed to two separate MLPs to regress $\tilde{\mathbf{z}}_{BG_l,pol}$ and $\tilde{\boldsymbol{\sigma}}_{BG_l,pol}^2$, respectively.}
  \label{fig:NNArch}
% \vspace{-3mm}
\end{figure}

The overall structure of the deep network is shown in Figure~\ref{fig:NNArch}.
First, the input image is processed by a Convolutional Neural Network (CNN), based on the shallow DroNet architecture \cite{loquercio2018dronet}.
The extracted features are then processed by two separate multilayer perceptrons (MLPs) that estimate the mean $\tilde{\mathbf{z}}_{BG_l,pol}$ and the variance $\tilde{\boldsymbol{\sigma}}_{BG_l,pol}^2$ of a multivariate normal distribution, respectively.
A similar network architecture for mean-variance estimation was proposed in~\cite{nix1994estimating}.

\subsubsection{Training Procedure}
We train the network in two stages.

In the first stage, the parameters of the CNN and $\text{MLP}_{\mathbf{z}}$, denoted by $\boldsymbol{\theta}_{\text{CNN}}$ and $\boldsymbol{\theta}_{\mathbf{z}}$, are jointly learned by minimizing a loss over groundtruth poses for images with visible {\waypoints}:
\begin{equation}
    \{\boldsymbol{\theta}_\text{CNN}^\ast, \boldsymbol{\theta}_{\tilde{\mathbf{z}}_i}^\ast\} = \arg \min_{\boldsymbol{\theta}_\text{CNN}, \boldsymbol{\theta}_{\tilde{\mathbf{z}}_i}} \sum_{i=1}^{N} || \mathbf{y}_i - \tilde{\mathbf{z}}_i ||_2^2,
\end{equation}
where $\mathbf{y}_i$ denotes the groundtruth pose and $N$ denotes the dataset size.

In the second stage, the training set is extended to also include images that do not show visible {\waypoints}. In this stage only the parameters $\boldsymbol{\theta}_{\sigma^2}$ of the subnetwork $\text{MLP}_{\sigma^2}$ are trained, while keeping the other weights fixed. We minimize the loss function proposed by \cite{nix1994estimating}, which amounts to the negative log-likelihood of a multivariate normal distribution with uncorrelated covariance:
\begin{equation}
    - \log p\left(\mathbf{y} \mid \tilde{\mathbf{z}}_i, \tilde{\boldsymbol{\sigma}}^2 \right) \propto \sum_{j=1}^{4} \log \tilde{\sigma}_j^2 + \frac{\left(y_j - \tilde{z}_j \right)^2}{\tilde{\sigma}_j^2} .
\end{equation}

Our use of mean-variance estimation is motivated by studies that have shown that it is a computationally efficient way to obtain uncertainty estimates~\cite{khosravi2011comprehensive}.

\subsubsection{Training Data Generation}
We collect a set of images from the forward-facing camera on the drone and associate each image with the relative pose of the {\waypoint} with respect to the body frame of the quadrotor.
In real-world experiments, we use the quadrotor and leverage the onboard state estimation pipeline to generate training data. 
The platform is initialized at a known position relative to a {\waypoint} and subsequently carried through the environment while collecting images and corresponding relative {\waypoint} poses.  
To collect training data, it is not necessary to have complete tracks available. A single {\waypoint} placed in different environments suffices, as the perception system only needs to estimate the relative pose with respect to the next {\waypoint} at test time.
Moreover, in contrast to Kaufmann et al.~\cite{Kaufmann18arxiv}, the perception system is never trained on data from tracks and environments it is later deployed in.

\begin{figure}
    \centering
    \includegraphics[width=\linewidth, trim={100, 100, 100, 100}, clip]{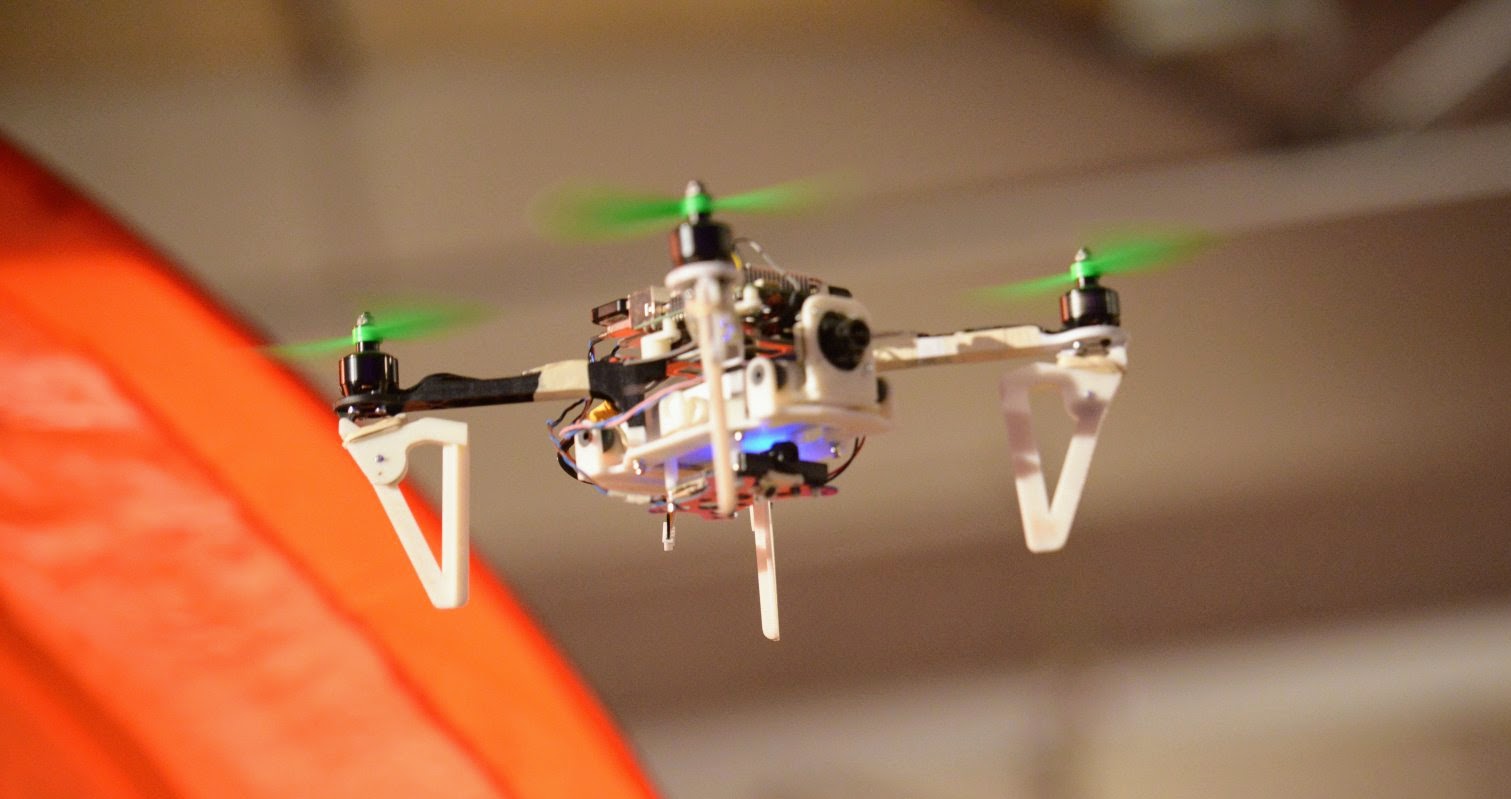}
    \caption{Our platform, equipped with an Intel UpBoard and a Qualcomm Snapdragon Flight.}
   \label{fig:platform}
%   \vspace{-3mm}
\end{figure}

\subsection{Mapping System}
The mapping system takes as input a measurement from the perception system and outputs a filtered estimate of the current track layout.
By correcting the gates with the measurements from the CNN, {\waypoint} displacement and accumulated VIO drift can be compensated for.
The mapping part of our pipeline can be divided into two stages: measurement assignment stage and filter stage.
\subsubsection{Measurement Assignment}
We maintain a map of all {\waypoints} $l=1 ... N_l$ with states $\hat{\mathbf{x}}_{OG_l}=[\hat{\mathbf{t}}_{OG_l}, \hat{\phi}_{OG_l}]^\top$ corresponding to {\waypoint} translation $\hat{\mathbf{t}}_{OG_l}$ and yaw $\hat{\phi}_{OG_l}$ with respect to the odometry frame $O$.
The output of the perception system is used to update the pose $\hat{\mathbf{x}}_{OG_l}$ of the next {\waypoint} to be passed. 
To assign a measurement to a {\waypoint}, the measurement is transformed into the odometry frame and assigned to the closest {\waypoint}.
If a measurement is assigned to a {\waypoint} that is not the next {\waypoint} to be passed, it is discarded as an outlier. 
We keep track of the next {\waypoint} by detecting {\waypoint} traversals.
The detection of a {\waypoint} traversal is done by expressing the quadrotor's current position in a {\waypoint}-based coordinate frame. 
In this frame, the condition for traversal can be expressed as
\begin{equation}
    ~_{G_l}\hat{\mathbf{t}}_{G_lB,x} \geq 0.
\end{equation}

\subsubsection{Extended Kalman Filter}
The prediction of the network in body frame $B$ is given by $\tilde{\mathbf{z}}_{BG,pol}=[\tilde{r}, \tilde{\theta}, \tilde{\psi}, \tilde{\phi}]^\top$ containing the spherical coordinates $[\tilde{r}, \tilde{\theta}, \tilde{\psi}]^\top$ and yaw $\tilde{\phi}$ of the {\waypoint}, and the corresponding variance $\tilde{\boldsymbol{\sigma}}_{BG,pol}^2$. The transformation into the Cartesian representation $\tilde{\mathbf{z}}_{BG}$ leads to
\begin{align}
\tilde{\mathbf{z}}_{BG}&= \mathbf{f}(\tilde{\mathbf{z}}_{BG,pol})
= \begin{bmatrix} \tilde{r} \sin{\tilde{\theta}} \cos{\tilde{\psi}}  \\ \tilde{r} \sin{\tilde{\theta}} \sin{\tilde{\psi}} \\ \tilde{r} \cos{\tilde{\theta}} \\ \tilde{\phi} \end{bmatrix} \\
\tilde{\boldsymbol{\Sigma}}_{BG}&= \mathbf{J}_{\mathbf{f}}|_{\tilde{\mathbf{z}}_{pol}} \tilde{\boldsymbol{\Sigma}}_{BG,pol} \mathbf{J}_{\mathbf{f}}^\top|_{\tilde{\mathbf{z}}_{pol}},
\end{align}
where $\mathbf{J}_{\mathbf{f},i,j}=\frac{\partial f_i}{\partial x_{pol,j}}$ is the Jacobian of the conversion function $\mathbf{f}$ and $\mathbf{J}_{\mathbf{f}}|_{\mathbf{z}_{pol}}$ is its evaluation at $\mathbf{z}_{pol}$.
To integrate neural network predictions reliably into a map with prior knowledge of the {\waypoints}, we represent each {\waypoint} with its own EKF.
We treat the prediction $\tilde{\mathbf{z}}_{BG}$ and $\tilde{\boldsymbol{\Sigma}}_{BG}$ at each time step as a measurement and associated variance, respectively.
Similar to the state, $\tilde{\mathbf{z}}_{BG}=[\tilde{\mathbf{t}}_{BG}^\top, \tilde{\phi}_{BG}]^\top$ consists of a translation $\tilde{\mathbf{t}}_{BG}$ and rotation $\tilde{\phi}_{BG}$ around the world \mbox{$z$-axis}.
Since our measurement and states have different origin frames, we can formulate the EKF measurement as follows:
\begin{align}
\tilde{\mathbf{z}}_k &= \mathbf{H}_k \hat{\mathbf{x}}_k + \mathbf{w},~ \quad \mathbf{w}\sim\mathcal{N}\left(\boldsymbol{\mu}_k, \boldsymbol{\sigma}_k\right) \\
\mathbb{E} [\tilde{\mathbf{z}}_k] &= \begin{bmatrix} \mathbf{R}_{OB,k}^{-1} ~_O\mathbf{t}_{OG,k} - \mathbf{R}_{OB,k}^{-1} ~_O\mathbf{t}_{OB,k} \\ \phi_{OG,k} - \phi_{OB,k} \end{bmatrix}. \nonumber
\end{align}
Now with $\hat{\mathbf{x}}_k = [ _O\mathbf{t}_{OG,k}, \quad \phi_{OG,k}]^\top$ we can write
\begin{align}
\mathbf{H}_k &= \begin{bmatrix} \mathbf{R}_{OB,k}^{-1} & \mathbf{0} \\ \mathbf{0} & 1 \end{bmatrix}
\label{eq:kalman_hmatrix} \\
\boldsymbol{\mu}_k &= \begin{bmatrix} -\mathbf{R}_{OB,k}^{-1} ~_O\mathbf{t}_{OB,k} \\ -\phi_{OB,k} \end{bmatrix} &
\boldsymbol{\Sigma}_k &= \tilde{\boldsymbol{\Sigma}}_{BG,k}
\label{eq:kalman_noisedistribution}
\end{align}
and, due to identity process dynamics and process covariance $\boldsymbol{\Sigma}_Q$, our prediction step becomes
\begin{align}
\hat{\mathbf{x}}_{k+1}^* &= \hat{\mathbf{x}}_{k} &
\hat{\mathbf{P}}_{k+1}^* &= \hat{\mathbf{P}}_k + \boldsymbol{\Sigma}_Q.
\label{eq:kalman_apriori}
\end{align}
The a-posteriori filter update can be summarized as follows:
\begin{equation}
\begin{split}
\mathbf{K}_k &= \hat{\mathbf{P}}_k^* \mathbf{H}_k \left( \tilde{\boldsymbol{\Sigma}}_{BG,k} + \mathbf{H}_k \hat{\mathbf{P}}_k^* \mathbf{H}_k^\top \right)^{-1}  \\
\hat{\mathbf{x}}_{k+1} &= \hat{\mathbf{x}}_k^* + \mathbf{K}_k (\tilde{\mathbf{z}}_k - \boldsymbol{\mu}_k - \mathbf{H}_k \hat{\mathbf{x}}_k^*) \\
\hat{\mathbf{P}}_{k+1} &= \left( \mathbf{I} - \mathbf{K}_k \mathbf{H}_k \right) \hat{\mathbf{P}}_k^* \left( \mathbf{I} - \mathbf{K}_k \mathbf{H}_k \right)^\top + \mathbf{K}_k \tilde{\boldsymbol{\Sigma}}_{BG,k} \mathbf{K}_k^\top
\label{eq:kalman_aposteriori}
\end{split}
\end{equation}
with $\hat{\mathbf{P}}_k$ as the estimated covariance and the superscript $^*$ indicating the a-priori predictions.

\subsection{Planning and Control System}
The planning and control stage is split into two asynchronous modules. First, low-level waypoints are generated from the estimated {\waypoint} position and a desired path
is generated by linearly interpolating between the low-level waypoints. Second, locally feasible control trajectories are planned and tracked using a model predictive control scheme.

\subsubsection{Waypoint Generation}
\label{sec:waypoint_generation}
For each {\waypoint} in our map we generate two waypoints: one lying in front of the {\waypoint} relative to the current quadrotor position and one lying after the {\waypoint}.
Both waypoints are set with a positive and negative offset $\mathbf{p}_{wp,l\pm}$ in the \mbox{$x$ direction} with respect to the {\waypoint}~$l$:
\begin{equation}
\mathbf{p}_{wp,l\pm} = _O\mathbf{t}_{OG_l} + \mathbf{R}_{OG_l} [\pm x_G, 0, 0]^\top,
\end{equation}
where $x_G$ is a user-defined constant accounting for the spatial dimension of {\waypoint}~$l$.
We then linearly interpolate a path from waypoint to waypoint and use it as a reference for our controller.

\begin{figure*}[!t]
\centering
\hspace{-5pt}
\subfloat[\label{fig:plot_track1} Track 1, $v_{max}=\SI{2}{\meter\per\second}$, $\rho=\SI{2}{\meter}$]{
    \includegraphics[width=0.31\linewidth, trim={15 0 30 20},clip]{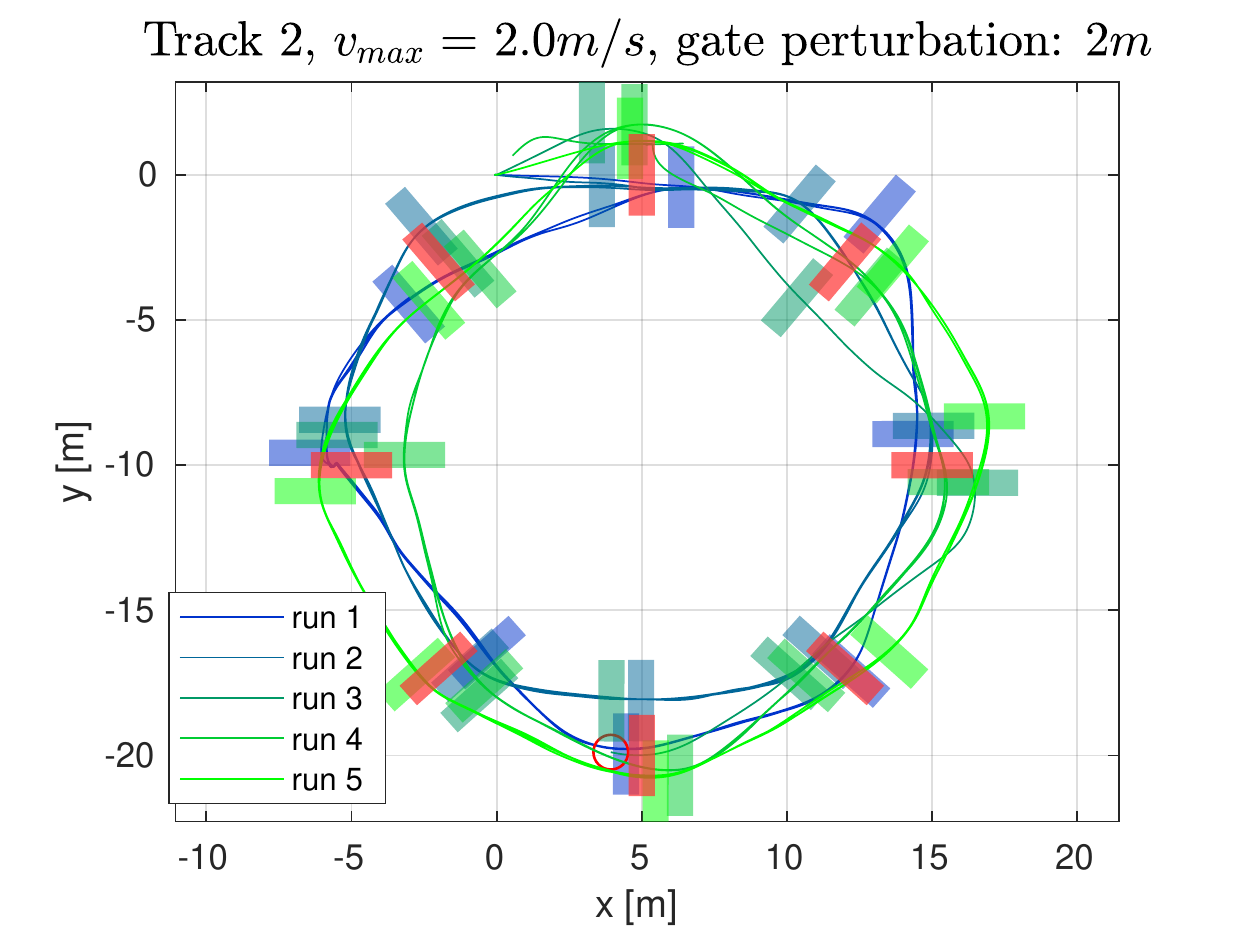}}
\hspace{4pt}
\subfloat[\label{fig:plot_track2} Track 2, $v_{max}=\SI{2}{\meter\per\second}$, $\rho=\SI{2}{\meter}$]{
    \includegraphics[width=0.31\linewidth, trim={15 0 30 20},clip]{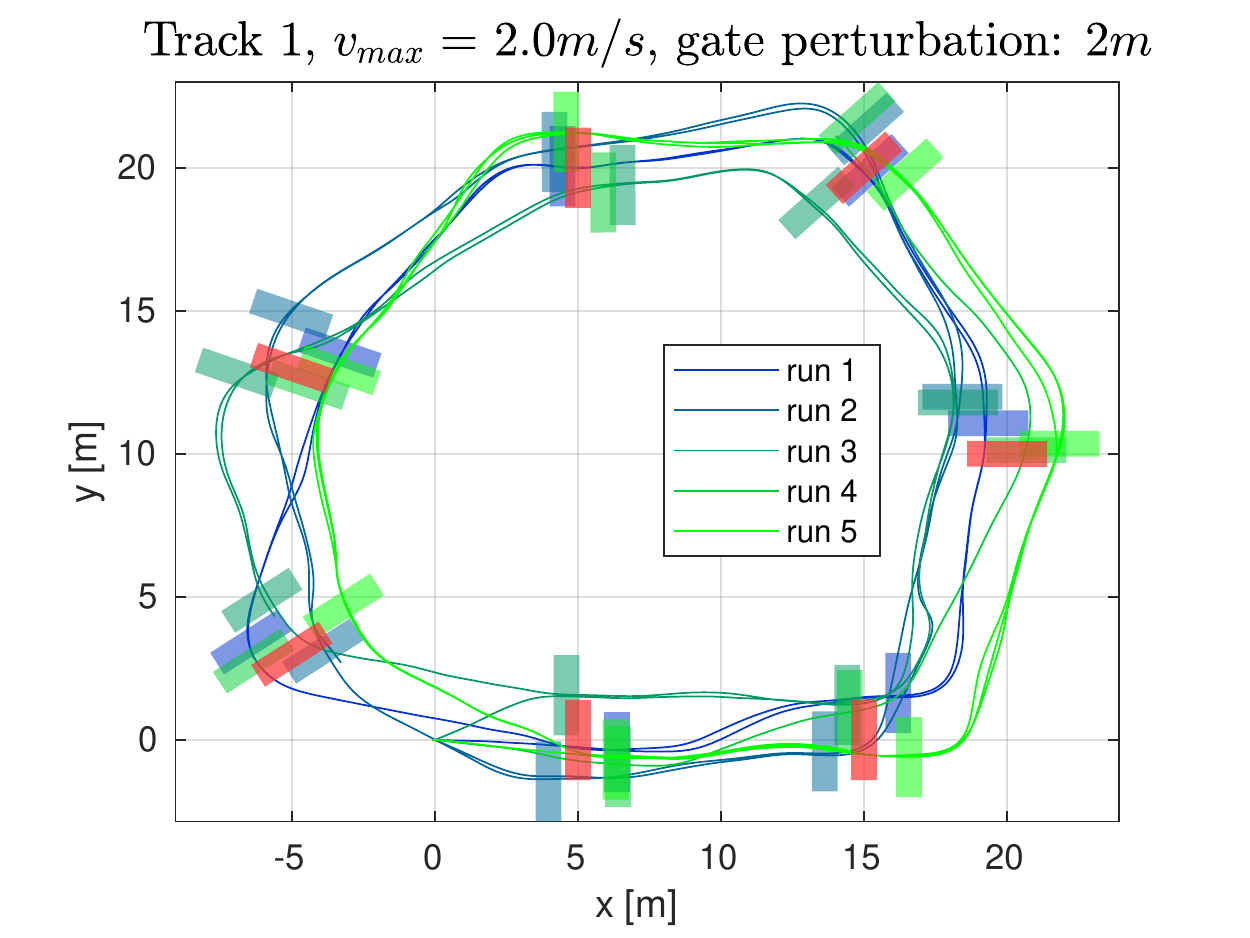}}
\hspace{4pt}
\subfloat[\label{fig:plot_track3} Track 3, $v_{max}=\SI{2}{\meter\per\second}$, $\rho=\SI{1}{\meter}$]{
    \includegraphics[width=0.31\linewidth, trim={15 0 30 20},clip]{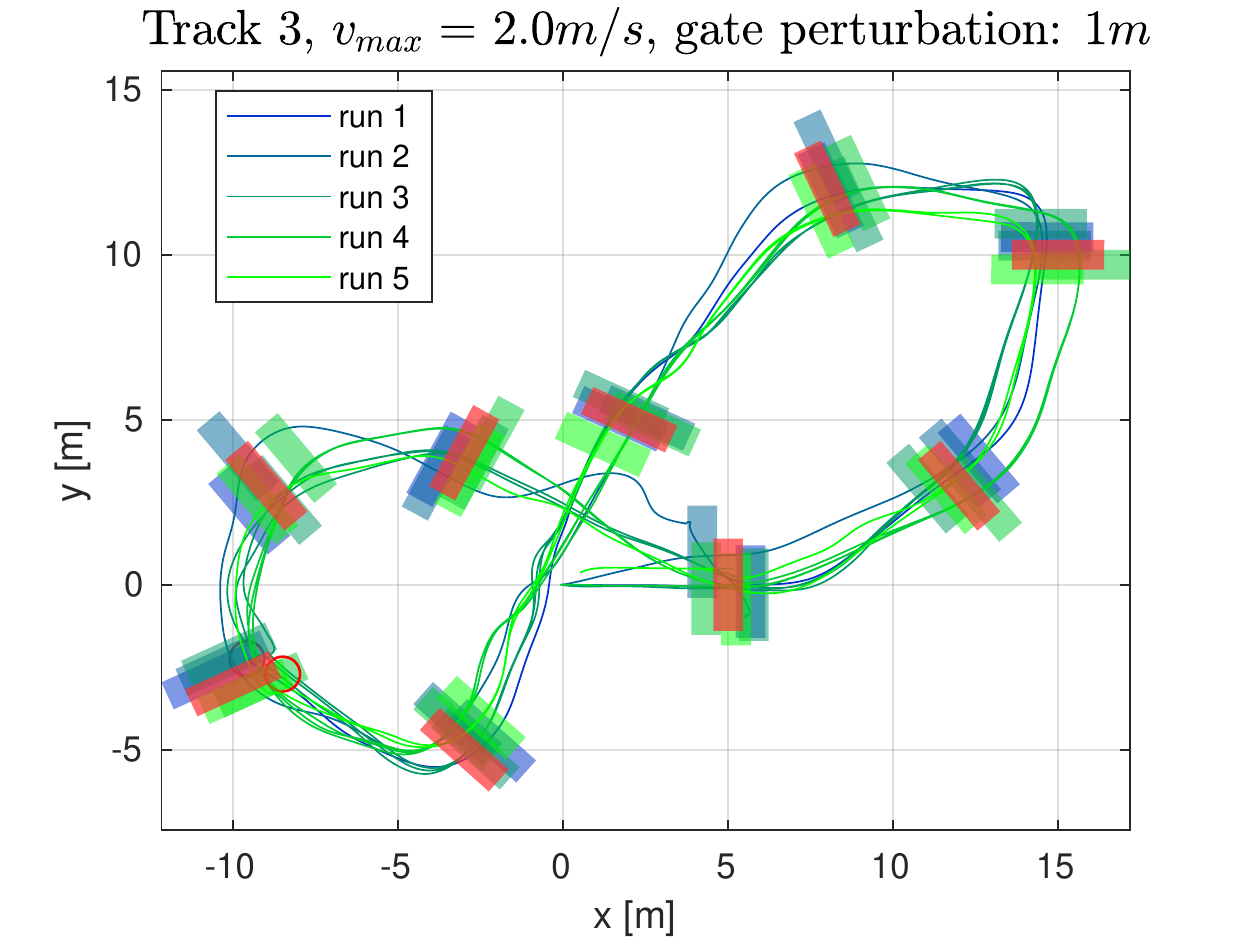}}
\vspace{-5pt}
\subfloat[\label{fig:overview_track1} Ours on track 1]{
    \includegraphics[width=0.31\linewidth, trim={0 5 15 20},clip]{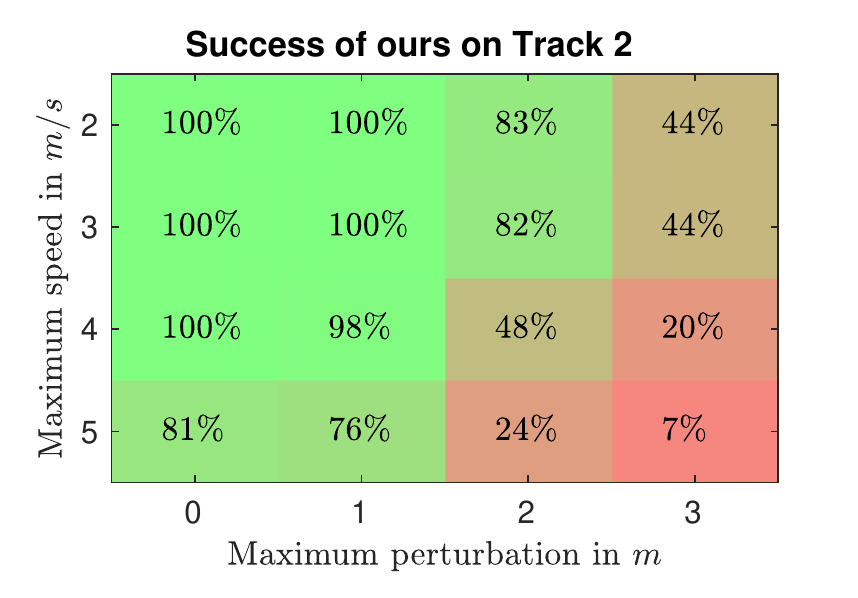}}
\hspace{5pt}
\subfloat[\label{fig:overview_track2} Ours on track 2]{
    \includegraphics[width=0.31\linewidth, trim={0 5 15 20},clip]{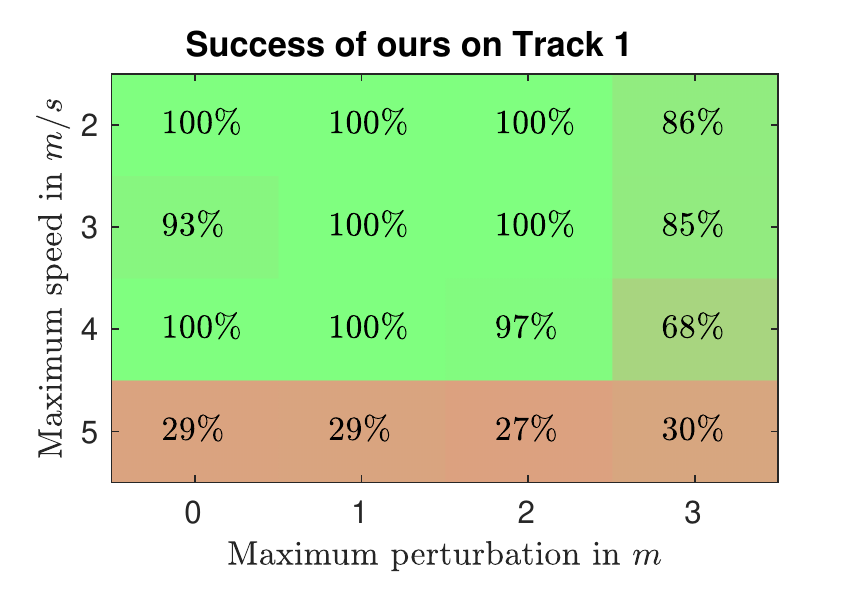}}
\hspace{5pt}
\subfloat[\label{fig:overview_track3} Ours on track 3]{
    \includegraphics[width=0.31\linewidth, trim={0 5 15 20},clip]{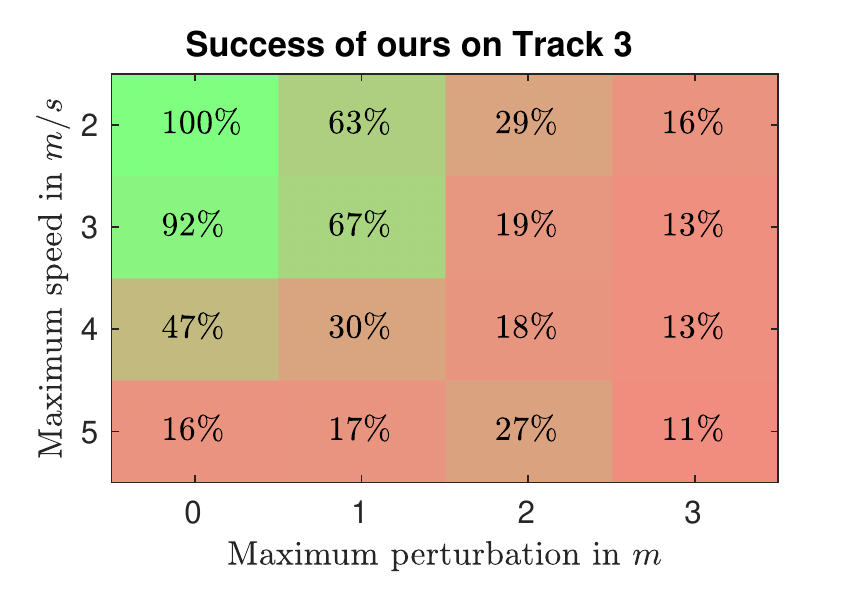}}
%\bigskip
\vspace{-5pt}
\subfloat[\label{fig:overview_baseline1} Baseline on track 1]{
    \includegraphics[width=0.31\linewidth, trim={0 5 15 20},clip]{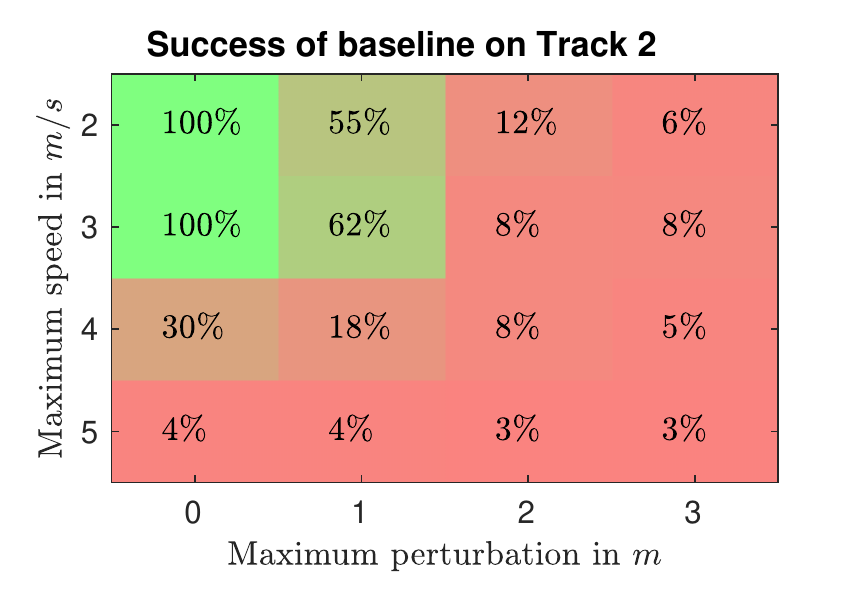}}
\hspace{5pt}
\subfloat[\label{fig:overview_baseline2} Baseline on track 2]{
    \includegraphics[width=0.31\linewidth, trim={0 5 15 20},clip]{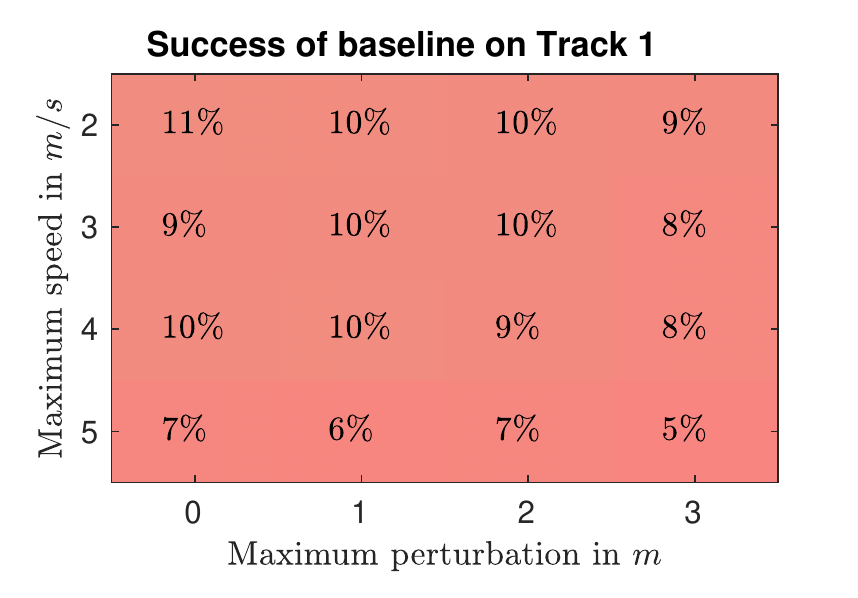}}
\hspace{5pt}
\subfloat[\label{fig:overview_baseline3} Baseline on track 3]{
    \includegraphics[width=0.31\linewidth, trim={0 5 15 20},clip]{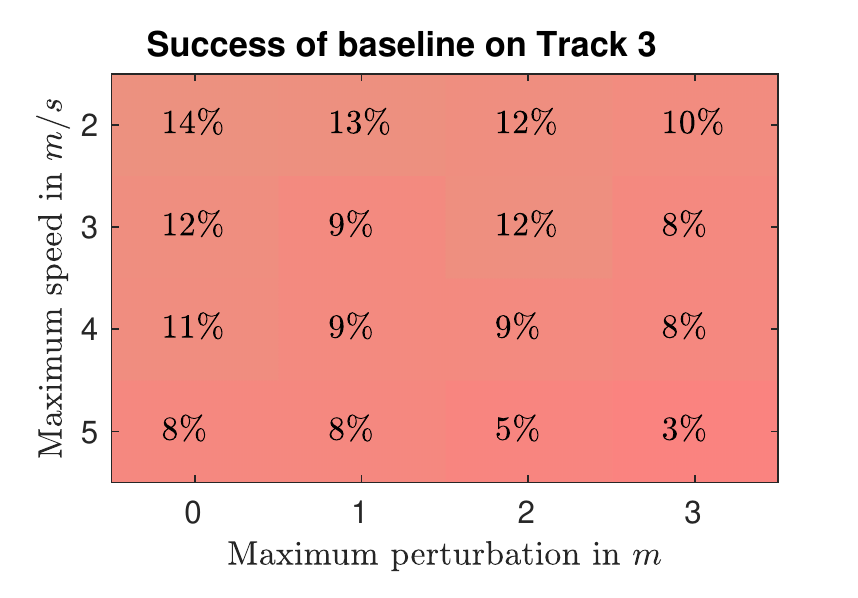}}
    
\caption{Results of the simulation experiments. We compare the presented approach to the baseline~\cite{Jung18ral} on three tracks, at different speeds and track perturbations. \textbf{(a)-(c):}~Perturbed tracks and example trajectories flown by our approach. \textbf{(d)-(f):}~Success rate of our method. For each data point, 5 experiments were performed with random initial {\waypoint} perturbation. \textbf{(g)-(i):}~Success rate of the baseline method.}
\label{fig:simulation_success}
\vspace{-3mm}
\end{figure*}

\subsubsection{Model Predictive Control}
\label{sec:mpc}
We formulate the control problem as a quadratic optimization problem which we solve using sequential quadratic programming as described in \cite{Falanga18iros}:
\begin{gather}
\min_{\mathbf{u}} \int_{t_0}^{t_f} \left( \bar{\mathbf{x}}_t^\top(t) \mathbf{Q} \bar{\mathbf{x}}_t(t) + \bar{\mathbf{u}}_t^\top(t) \mathbf{R} \bar{\mathbf{u}}_t(t) \right)dt \nonumber \\
\begin{aligned}
\bar{\mathbf{x}}(t) = \mathbf{x}(t) &- \mathbf{x}_r(t) & \bar{\mathbf{u}}(t) = \mathbf{u}(t) &- \mathbf{u}_r(t) \\
\text{subject to} \quad
\mathbf{r}(\mathbf{x}, \mathbf{u}) &= 0 & \mathbf{h}(\mathbf{x}, \mathbf{u}) &\leq 0. \nonumber
\label{eq:mpc_formulation}
\end{aligned}
\end{gather}
The states $\mathbf{x}$ and inputs $\mathbf{u}$ are weighted with positive diagonal matrices $\mathbf{Q}$ and $\mathbf{R}$ with respect to a reference $\mathbf{x}_r$ and $\mathbf{u}_r$.
The equality and inequality constraints, $\mathbf{r}$ and $\mathbf{h}$ respectively, are used to incorporate the vehicle dynamics and input saturations.
The reference is our linearly sampled path along which the MPC finds a feasible trajectory.
Note that we can run the control loop independent of the detection and mapping pipeline and reactively stabilize the vehicle along the changing waypoints.

\section{Experimental Setup}\label{sec:experiments}
We evaluate the presented approach in simulation and on a physical system.
%
%\subsection{Experimental Setup}

\subsection{Simulation}
%\label{sec:simulation_experiments}
We use RotorS~\cite{Furrer16} and Gazebo~\cite{koenig2004gazebo}
%\footnote{http://gazebosim.org/}
for all simulation experiments.
To train the perception system, we generated 45,000 training images by randomly sampling camera and {\waypoint} positions and computing their relative poses. 
For quantitative evaluation, a $100\%$ successful trial is defined as completing 3 consecutive laps without crashing or missing a {\waypoint}.
If the MAV crashes or misses a {\waypoint} before completing $3$ laps, the success rate is measured as a fraction of completed {\waypoints} out of $3$ laps: for instance, completing $1$ lap counts as $33.3\%$ success.

\subsection{Physical System}
%\label{sec:realworld_experiments}
In all real-world experiments and data collection we use an in-house MAV platform with an Intel UpBoard
%\footnote{https://www.up-board.org/up/}
as the main computer running the CNN, EKF, and MPC. Additionally we use a Qualcomm Snapdragon Flight
%\footnote{https://developer.qualcomm.com/hardware/qualcomm-flight}
as a visual-inertial odometry unit.
The platform is shown in Fig.~\ref{fig:platform}.
The CNN reaches an inference rate of $\sim \SI{10}{\Hz}$ while the MPC runs at $\SI{100}{\Hz}$.
With a take-off weight of $\SI{950}{\gram}$ the platform reaches thrust-to-weight ratio of $\sim 3$.

We collect training data for the perception system in five different environments, both indoors and outdoors.
Example images from the environments are shown in Fig.~\ref{fig:training_data}.
In total, we collected 32,000 images.

\section{Results}

Results are shown in the supplementary video at\\ \videolink.

\subsection{Simulation}
We first present experiments in a controlled, simulated environment. 
The aim of these experiments is to thoroughly evaluate the presented approach both quantitatively and qualitatively and compare it to a baseline~-- the method of Jung et al.~\cite{Jung18ral}.
The baseline was trained on the same data as our approach.

We evaluate the two methods on three tracks of increasing difficulty.
Figs.~\ref{fig:plot_track1}-\subref*{fig:plot_track3} show an illustration of the three race tracks
and plot the executed trajectories together with the nominal {\waypoint} positions in red and the actual displaced {\waypoint} positions in the corresponding track color.
Our approach achieved successful runs in all environments, with speeds up to $\SI{4}{\meter\per\second}$ in the first two tracks.
Additionally, {\waypoint} displacement was handled robustly up to a magnitude of $\SI{2}{\meter}$ before a significant drop in performance occurred.
\mbox{Figs.~\ref{fig:overview_track1}-\subref*{fig:overview_baseline3}} show the success rate of our method and the baseline on the three tracks, under varying speed and track perturbations.
Our approach outperforms the baseline by a large margin in all scenarios.
This is mainly because the baseline relies on the permanent visibility of the next {\waypoint}.
Therefore, it only manages to complete a lap in the simplest first track where the next gate can always be seen.
In the more complex second and third tracks, the baseline passes at most one or two gates.
In contrast, due to the integration of prior information from demonstration and approximate mapping, our approach is successful on all tracks, including the very challenging third one.

\begin{figure}
    \centering
    \includegraphics[width=\linewidth,trim={10, 0, 10, 10},clip]{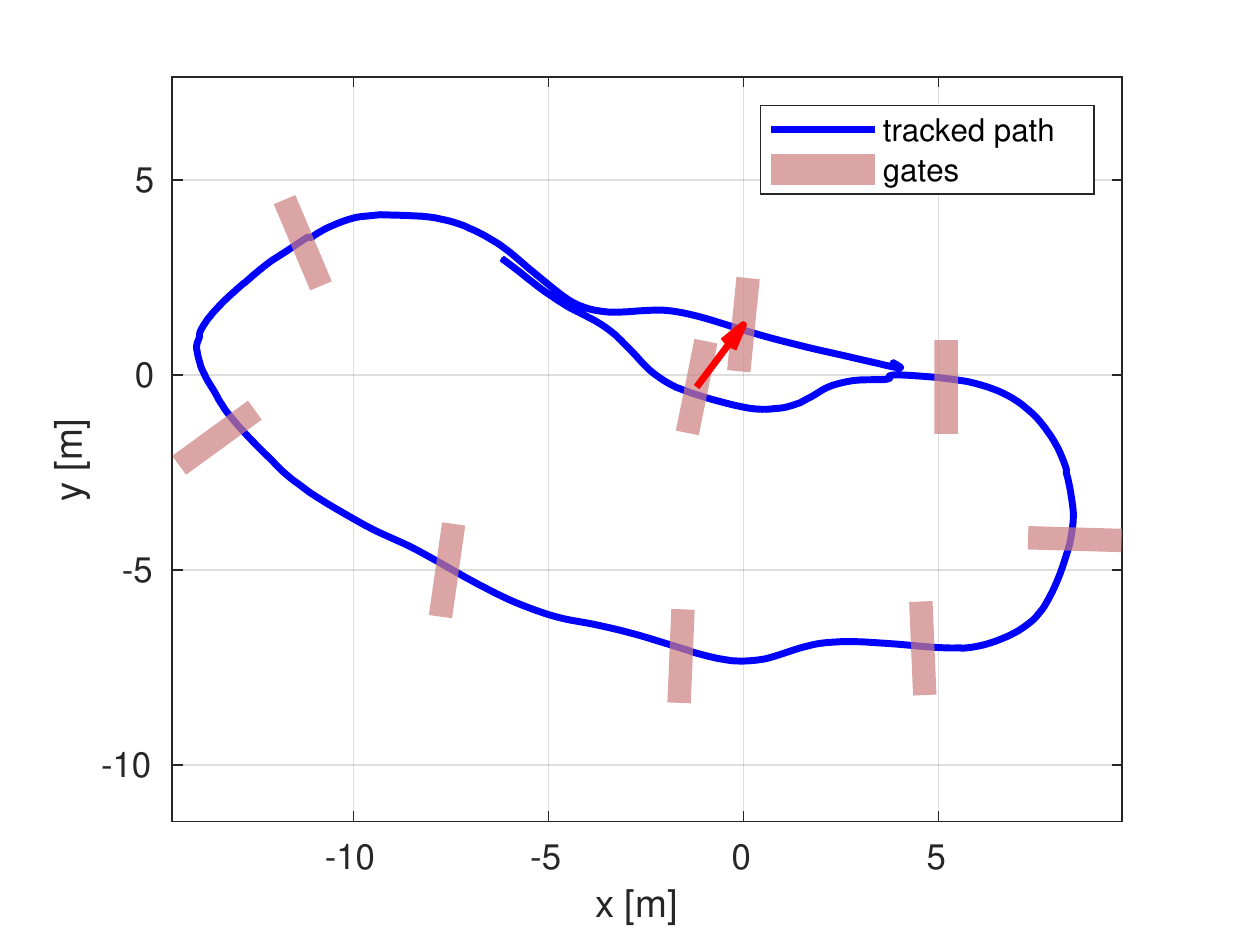}
    \caption{Trajectory flown through multiple {\waypoints}, one of which was moved as indicated by the red arrow. For visualization, only a single lap is illustrated.
    }
    \label{fig:plot_real_displacement}
%    \vspace{-3mm}
\end{figure}
% \vfill

\subsection{Physical System}
To show the capabilities of our approach on a physical platform, we evaluated it on a real-world track with 8 gates and a total length of 80 meters, shown in Fig.~\ref{fig:plot_real_displacement}. 
No training data for the perception system was collected in this environment.
\begin{figure}
    \centering
    \includegraphics{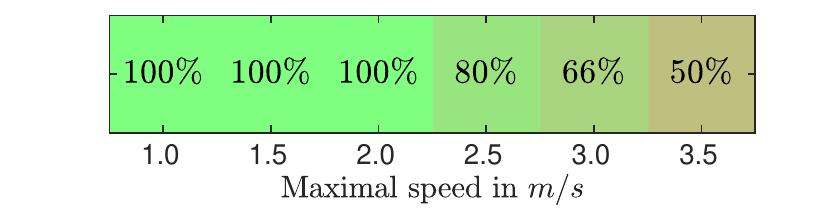}
    \caption{Success rates of our approach in the real-world experiment. The reader is encouraged to watch the supplementary video to see the presented approach in action.}
    \label{fig:table_real_success}
%    \vspace{-3mm}
\end{figure}
Fig.~\ref{fig:table_real_success} summarizes the results.
As in the simulation experiments, we measure the performance with respect to the average MAV speed.
As before, a success rate of $100\%$ requires 3 completed laps without crashing or missing a gate. 
Our approach confidently completed $3$ laps with speeds up to $\SI{2}{\meter\per\second}$ and managed to complete the track with speeds up to $\SI{3.5}{\meter\per\second}$. In contrast, the reactive baseline was not able to complete the full track even at $\SI{1.0}{\meter\per\second}$ (not shown in the figure).

An example recorded trajectory of our approach is shown in Fig.~\ref{fig:plot_real_displacement}.
Note that one of the gates was moved during the experiment, but our approach was robust to this change in the environment.
Our approach could handle {\waypoint} displacements of up to $\SI{3.0}{\meter}$ and complete the full track without crashing. 
The reader is encouraged to watch the supplementary video for more qualitative results on real tracks.
%
%These help understanding the level of agility the approach demonstrates.
%
\section{Conclusion}
We presented an approach to autonomous vision-based drone navigation. The approach combines learning methods and optimal filtering. 
In addition to predicting relative {\waypoint} poses, our network also estimates the uncertainty of its predictions.
This allows us to integrate the network outputs with prior information via an extended Kalman filter.

We showed successful navigation through both simulated and real-world race tracks with increased robustness and speed compared to a state-of-the-art baseline. 
The presented approach reliably handles gate displacements of up to $\SI{2}{\meter}$.
In the physical track, we reached speeds of up to $\SI{3.5}{\meter\per\second}$, outpacing the baseline by a large margin.

Our approach is capable of flying a new track with an approximate map obtained from a single demonstration flight.
This approach was used to win the IROS 2018 Autonomous Drone Race Competition,
where it outraced the second-placing entry by a factor of two.
%
%To further increase robustness, more training data of corner cases, such as multiple visible gates, could be used.
%Higher inference rates could allow to increase the maximum flight speed.
% \clearpage
% \balance
%\bibliographystyle{ieee}
%\bibliography{references}

\begin{thebibliography}{10}\itemsep=-1pt

\bibitem{Blosch15iros}
M.~Bloesch, S.~Omari, M.~Hutter, and R.~Siegwart.
\newblock Robust visual inertial odometry using a direct {EKF}-based approach.
\newblock In {\em IEEE/RSJ Int. Conf. Intell. Robot. Syst. (IROS)}, 2015.

\bibitem{drews2017aggressive}
P.~Drews, G.~Williams, B.~Goldfain, E.~A. Theodorou, and J.~M. Rehg.
\newblock Aggressive deep driving: Combining convolutional neural networks and
  model predictive control.
\newblock In {\em Conference on Robot Learning}, 2017.

\bibitem{Falanga18iros}
D.~Falanga, P.~Foehn, P.~Lu, and D.~Scaramuzza.
\newblock {PAMPC}: Perception-aware model predictive control for quadrotors.
\newblock In {\em IEEE/RSJ Int. Conf. Intell. Robot. Syst. (IROS)}, 2018.

\bibitem{fehr2018visual}
M.~Fehr, T.~Schneider, M.~Dymczyk, J.~Sturm, and R.~Siegwart.
\newblock Visual-inertial teach and repeat for aerial inspection.
\newblock {\em arXiv:1803.09650}, 2018.

\bibitem{Forster17troSVO}
C.~Forster, Z.~Zhang, M.~Gassner, M.~Werlberger, and D.~Scaramuzza.
\newblock {SVO}: Semidirect visual odometry for monocular and multicamera
  systems.
\newblock {\em {IEEE} Trans. Robot.}, 33(2), 2017.

\bibitem{Furrer16}
F.~Furrer, M.~Burri, M.~Achtelik, and R.~Siegwart.
\newblock {RotorS}{\textemdash}{A} modular {Gazebo} {MAV} simulator framework.
\newblock In {\em Robot Operating System (ROS)}. Springer, Cham, 2016.

\bibitem{Giusti16ral}
A.~Giusti, J.~Guzzi, D.~C. Cire{\c{s}}an, F.-L. He, J.~P. Rodr{\'i}guez,
  F.~Fontana, M.~Faessler, C.~Forster, J.~Schmidhuber, G.~D. Caro,
  D.~Scaramuzza, and L.~M. Gambardella.
\newblock A machine learning approach to visual perception of forest trails for
  mobile robots.
\newblock {\em {IEEE} Robot. Autom. Lett.}, 1(2), 2016.

\bibitem{Jung18ral}
S.~Jung, S.~Hwang, H.~Shin, and D.~H. Shim.
\newblock Perception, guidance, and navigation for indoor autonomous drone
  racing using deep learning.
\newblock {\em {IEEE} Robot. Autom. Lett.}, 3(3), 2018.

\bibitem{Kaufmann18arxiv}
E.~Kaufmann, A.~Loquercio, R.~Ranftl, A.~Dosovitskiy, V.~Koltun, and
  D.~Scaramuzza.
\newblock Deep drone racing: Learning agile flight in dynamic environments.
\newblock In {\em Conference on Robot Learning}, 2018.

\bibitem{khosravi2011comprehensive}
A.~Khosravi, S.~Nahavandi, D.~Creighton, and A.~F. Atiya.
\newblock Comprehensive review of neural network-based prediction intervals and
  new advances.
\newblock {\em IEEE Transactions on Neural Networks}, 22(9), 2011.

\bibitem{kim2015deep}
D.~K. Kim and T.~Chen.
\newblock Deep neural network for real-time autonomous indoor navigation.
\newblock {\em arXiv:1511.04668}, 2015.

\bibitem{koenig2004gazebo}
N.~Koenig and A.~Howard.
\newblock Design and use paradigms for {G}azebo, an open-source multi-robot
  simulator.
\newblock In {\em IEEE/RSJ Int. Conf. Intell. Robot. Syst. (IROS)}, 2014.

\bibitem{leutenegger2015keyframe}
S.~Leutenegger, S.~Lynen, M.~Bosse, R.~Siegwart, and P.~Furgale.
\newblock Keyframe-based visual-inertial odometry using nonlinear optimization.
\newblock {\em Int. J. Robot. Research}, 34(3), 2015.

\bibitem{Loianno17ral}
G.~Loianno, C.~Brunner, G.~McGrath, and V.~Kumar.
\newblock Estimation, control, and planning for aggressive flight with a small
  quadrotor with a single camera and {IMU}.
\newblock {\em {IEEE} Robot. Autom. Lett.}, 2(2), 2017.

\bibitem{loquercio2018dronet}
A.~Loquercio, A.~I. Maqueda, C.~R. del Blanco, and D.~Scaramuzza.
\newblock Dronet: Learning to fly by driving.
\newblock {\em {IEEE} Robot. Autom. Lett.}, 3(2), 2018.

\bibitem{Mellinger11icra}
D.~Mellinger and V.~Kumar.
\newblock Minimum snap trajectory generation and control for quadrotors.
\newblock In {\em {IEEE} Int. Conf. Robot. Autom. (ICRA)}, 2011.

\bibitem{Moon16iros}
H.~Moon, Y.~Sun, J.~Baltes, and S.~J. Kim.
\newblock The {IROS} 2016 competitions.
\newblock {\em {IEEE} Robot. Autom. Mag.}, 24(1), 2016.

\bibitem{morrell2018}
B.~Morrell, R.~Thakker, G.~Merewether, R.~G. Reid, M.~Rigter, T.~Tzanetos, and
  G.~Chamitoff.
\newblock Comparison of trajectory optimization algorithms for high-speed
  quadrotor flight near obstacles.
\newblock {\em {IEEE} Robot. Autom. Lett.}, 3(4), 2018.

\bibitem{mueller2017teaching}
M.~M{\"{u}}ller, V.~Casser, N.~Smith, D.~L. Michels, and B.~Ghanem.
\newblock Teaching {UAVs} to race using {UE4Sim}.
\newblock {\em arXiv:1708.05884}, 2017.

\bibitem{Mueller13icra}
M.~W. M{\"{u}}ller, M.~Hehn, and R.~D'Andrea.
\newblock A computationally efficient motion primitive for quadrocopter
  trajectory generation.
\newblock {\em {IEEE} Trans. Robot.}, 31(6), 2015.

\bibitem{MurArtal17tro}
R.~Mur-Artal and J.~D. Tard{\'o}s.
\newblock {ORB-SLAM2}: An open-source {SLAM} system for monocular, stereo, and
  {RGB-D} cameras.
\newblock {\em {IEEE} Trans. Robot.}, 33(5), 2017.

\bibitem{nix1994estimating}
D.~A. Nix and A.~S. Weigend.
\newblock Estimating the mean and variance of the target probability
  distribution.
\newblock In {\em IEEE International Conference On Neural Networks}, 1994.

\bibitem{schneider2018maplab}
T.~Schneider, M.~T. Dymczyk, M.~Fehr, K.~Egger, S.~Lynen, I.~Gilitschenski, and
  R.~Siegwart.
\newblock maplab: An open framework for research in visual-inertial mapping and
  localization.
\newblock {\em {IEEE} Robot. Autom. Lett.}, 3(3), 2018.

\bibitem{usenko2016direct}
V.~Usenko, J.~Engel, J.~St{\"u}ckler, and D.~Cremers.
\newblock Direct visual-inertial odometry with stereo cameras.
\newblock In {\em {IEEE} Int. Conf. Robot. Autom. (ICRA)}, 2016.

\end{thebibliography}

\end{document}